\documentclass[letterpaper, 10 pt, conference]{ieeeconf}

\IEEEoverridecommandlockouts                              
 
\usepackage{graphicx}
\usepackage{subfigure}
\usepackage{booktabs} 
\usepackage{algorithm}
\usepackage{algorithmic}
\usepackage{verbatim}
 
\usepackage{color,colortbl}
\usepackage{graphics} 
\usepackage{epsfig} 
\usepackage{times} 
\usepackage{amsmath} 
\usepackage{amssymb}  
\usepackage{hyperref}
\usepackage{mathrsfs}
\usepackage{amsfonts}
\usepackage{enumerate}
\usepackage{mathtools,amssymb}
\usepackage{makeidx}
\usepackage{ntheorem}
\theorembodyfont{\rmfamily}

\makeindex

\title{\LARGE \bf
	Safe Learning-based Tracking Control for Quadrotors under Wind Disturbances
}

\author{Lei Zheng, Rui Yang, Jiesen Pan, and Hui Cheng$^{*}$
	\thanks{L. Zheng is with the School of Electronics and Information Technology, Sun Yat-sen University, Guangzhou 510006, China.}%
	\thanks{R. Yang, J. Pan, H. Cheng are with the School of Computer Science and Engineering, Sun Yat-sen University, Guangzhou 510006, China.}%
	\thanks{*Corresponding author: chengh9@mail.sysu.edu.cn}%
}

\begin{document}
	\maketitle
	\thispagestyle{empty}
	\pagestyle{empty}

	\begin{abstract}
		Enforcing safety on precise trajectory tracking is critical for aerial robotics subject to wind disturbances. In this paper, we present a learning-based safety-preserving cascaded quadratic programming control (SPQC) for safe trajectory tracking under wind disturbances. The SPQC controller consists of a position-level controller and an attitude-level controller. Gaussian Processes (GPs) are utilized to estimate the uncertainties caused by wind disturbances, and then a nominal Lyapunov-based cascaded quadratic program (QP) controller is designed to track the reference trajectory. To avoid unexpected obstacles when tracking, safety constraints represented by control barrier functions (CBFs) are enforced on each nominal QP controller in a way of minimal modification. The performance of the proposed SPQC controller is illustrated through numerical validations of (a) trajectory tracking under different wind disturbances, and (b) trajectory tracking in a cluttered environment with a dense time-varying obstacle field under wind disturbances.
		
	\end{abstract}

	\section{Introduction}
	
	Safe trajectory tracking is an essential requirement for autonomous aerial vehicles, where an aerial vehicle is required to track a reference trajectory and avoid unexpected obstacles. To achieve high-accuracy tracking and avoid obstacles, the typical approach utilizes a high-level planner to generate a safe trajectory, and a low-level controller running at a much higher frequency than the planner to track the planned trajectory \cite{leung2020infusing}.
	
	In practice, for high-speed quadrotor tracking in a clutter scenario with dense obstacles and limited sensing range, it is challenging for a planner to replan a safe trajectory in real-time. On the other hand, it is difficult to obtain an accurate model for the low-level tracking controller due to wind disturbances, which may cause the safety-critical quadrotor system to deviate from the planned trajectory and even collide with obstacles. These challenges show an urgent need for designing a safe tracking controller that ensures both obstacle avoidance and accurate trajectory tracking for a quadrotor with a limited sensing range and under uncertain wind disturbances.
	
	In recent years, for safe trajectory tracking, the cascaded architecture~\cite{khan2020barrier,wu2016safety} using control barrier functions (CBFs) to ensure safety has shown great potential for addressing these challenges with practical feasibility. CBFs~\cite{ames2016control} have been widely used to enforce safety constraints, and have proved to be an effective strategy to avoid obstacles for safety-critical systems. In~\cite{Wang2018SafeLO}, a learning-based CBF is presented to constrain a quadrotor within a static ellipsoid safe region under wind disturbances. In~\cite{wu2016safety}, a cascaded quadratic program (QP) safety-critical controller incorporating CBFs and control Lyapunov functions (CLFs) is proposed. In this work, the quadrotor with a limited sensing range can asymptotically track a reference trajectory by constructing stability constraints (represented by CLFs), and simultaneously avoid obstacles by constructing safety constraints (represented by CBFs). Although this controller can handle time-varying constraints, a precise model of the system is required to enforce the constraints.
	\begin{figure}[!t]
		\centering
		\includegraphics[scale=0.32]{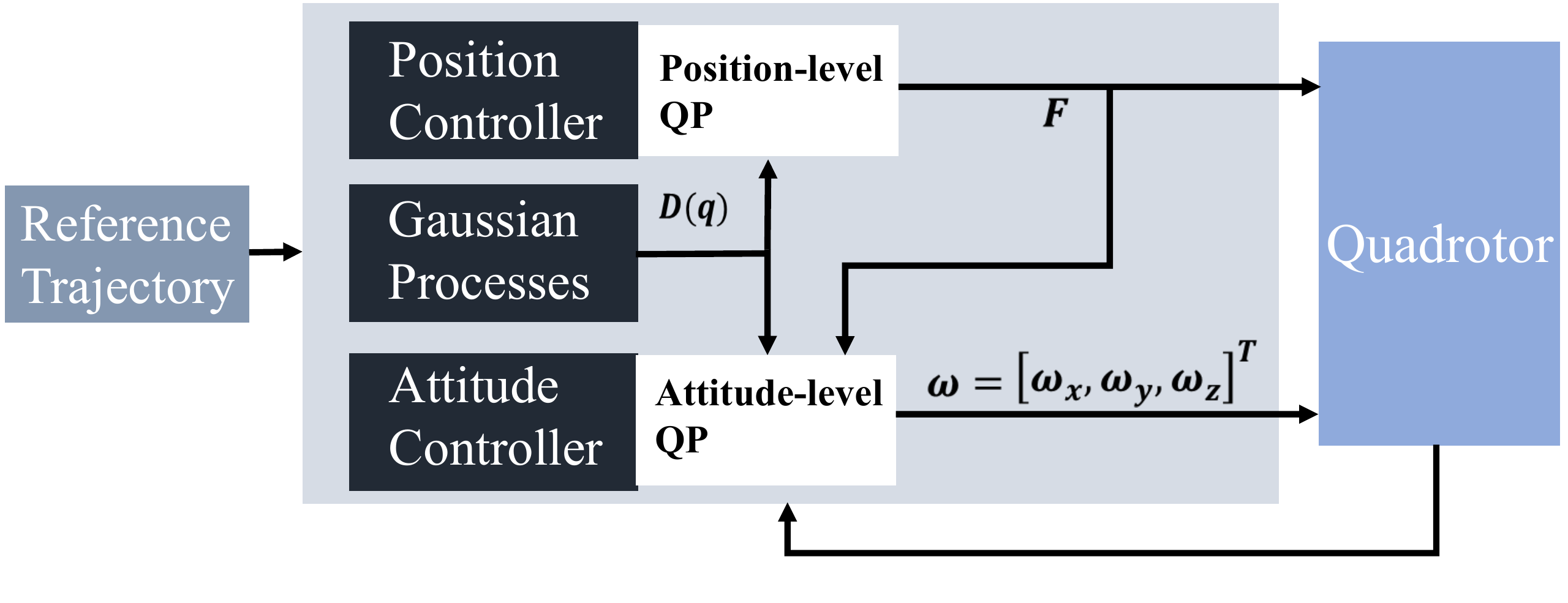}	\vspace{-0.2cm}
		\caption{{\bf{	
					The cascaded controller consists of a position-level QP and an attitude-level QP. Reference inputs are provided to position and attitude controllers. The GPs estimate the wind disturbances and generate a high confidence interval bound $D(q)$. With the $D(q)$, the position controller generates the desired thrust $F$, and the attitude controller generates the desired body rotational rates $\omega$.}}}
		\vspace{-0.6cm}
		\label{fig:cascaded controller}
	\end{figure}

    Alternatively, Model Predictive Control (MPC)~\cite{Mayne2000ConstrainedMP} is an effective finite-horizon optimal control method to handle constraints naturally in a safe tracking control problem.
    In\cite{castillo2018model}, a Nonlinear MPC is proposed to track the desired trajectory while avoiding obstacles. While this controller can mediate the trade-off between safety and tracking, disturbances are not considered in its model.~\cite{pereidaall} proposes a Nonlinear MPC for safe trajectory tracking with obstacle avoidance capacity under constant disturbances. Although this method can efficiently deal with obstacles when tracking a trajectory, the considered obstacles are assumed to be static.
	  
	Considering the limitations of current approaches, a safe tracking controller is desirable to be developed for a quadrotor subject to a limited sensing range to ensure accurate trajectory tracking and enforce safety constraints under wind disturbances. Particularly, it should drive the quadrotor back to the predefined reference trajectory after safely avoiding unexpected static and dynamic obstacles. 
	
	In this paper, drawing inspiration from the cascaded architecture~\cite{wu2016safety}, we propose a learning-based safety-preserving cascaded QP control (SPQC) approach composed of a position-level controller and an attitude-level controller, as illustrated in Fig.~\ref{fig:cascaded controller}. We utilize the Gaussian Process (GP) to estimate a high confidence interval $D(q)$ accounting for the model uncertainties caused by wind disturbances. Then, a local motion planning (LMP) method is designed based on the $D(q)$ to generate a reference orientation for the attitude controller. In each QP controller, a stability constraint based on CLF is formulated to develop a nominal Lyapunov-based cascaded tracking controller for accurate tracking. 
	Considering safety, the Iterative Regional Inflation by Semidefinite programming (IRIS) algorithm~\cite{deits2015computing} is adopted to construct a time-varying obstacle-free safe region and then use this safe region to design two CBFs to enforce safety constraints both on the position level and the attitude level for the nominal cascaded tracking controller. Finally, the safety constraints are enforced to minimally modify the Lyapunov-based tracking controller to formulate the SPQC controller.  
	
	The $\mathbf{main\ contributions}$ of this paper are presented as follows: 
	\begin{itemize}
		\item A learning-based SPQC controller consists of a position-level and an attitude-level controller is presented for quadrotors to safely track the reference trajectory with obstacle avoidance capacity under wind disturbances. An LMP algorithm is designed for the attitude-level controller for accurate tracking under wind disturbances.
		
		\item Two CBFs are designed to enforce safety constraints on the position and attitude level for the quadrotor to minimally modify the controls generated by the nominal tracking controller via constrained QPs.
		
		\item The proposed SPQC controller is validated through numerical simulations on a quadrotor in a cluttered environment under varying wind disturbances.
	\end{itemize}
	
	This paper is organized as follows: 
	Section~\ref{section:preliminaries} presents the preliminaries used in this paper. Section~\ref{section:SPQC} illustrates the proposed SPQC controller. Section~\ref{section:experiment} shows simulation results to validate and clarify our SPQC controller. Finally, conclusions  are drawn in Section~\ref{section:conclusion}.
	\vspace{-0.1cm}
	\section{Preliminaries}
	\label{section:preliminaries}
	\vspace{-0.12cm}
	In this section, we briefly provide a review of CBFs~\cite{ames2016control} and quadrotor dynamics to formulate the safe learning-based tracking control problem for Quadrotors. 
	\vspace{-0.18cm}
	\subsection{Control Barrier Function} 
	Consider a control affine system
	\vspace{-0.2cm}
	\begin{equation}
	\dot{x}=f(x)+G(x)u,
	\label{dynamics}	
	\vspace{-0.1cm}
	\end{equation}
	where $x\in\mathcal{X}\subseteq\mathbb{R}^{n}$, $u\in\mathcal{U}\subseteq\mathbb{R}^{m}$ denote the state and the control input of the system, respectively. The function $f : \mathbb{R}^{n}\rightarrow\mathbb{R}^{n}$ and $G : \mathbb{R}^{n}\rightarrow\mathbb{R}^{n \times m}$ are Lipschitz continuous.
	
	The $\mathit{safety\ set}\ \mathcal{S}$ of the system can be defined by
	\vspace{-0.2cm}
	\begin{equation}
	\mathcal{S}:=\{x\in\mathcal{X}|h(x)\geq0\},
	\label{safety set}
	\vspace{-0.1cm}
	\end{equation}
	where $h:\mathbb{R}^{n}\rightarrow\mathbb{R}$ is a continuously differentiable function related to state constraints.

	\noindent\textbf{Definition 2}\noindent\textbf{.}
	The set $\mathcal{S}$ is called $\mathit{forward\ invariant}$, if for every $x_{0}\in\mathcal{S}$, $x(t,x_{0})\in\mathcal{S}$ for all $t\in \mathbb{R}_{0}^{+}$.
	
	To ensure forward invariance of $\mathcal{S}$, e.g. quadrotors stay in the collision-free safety set at all times, we consider the following definition.
	
	\noindent\textbf{Definition 3} (Definition 5 of \cite{ames2016control})\noindent\textbf{.}
	For the dynamical system~(\ref{dynamics}), given a set $\mathcal{S}\subset{\mathbb{R}}^{n}$ defined by~(\ref{safety set}) for a continuously differentiable function $h:\mathbb{R}^{n}\rightarrow\mathbb{R}$, the function $h$ is called a $\mathit{Zeroing \ Control\  Barrier \  Function}$ $(ZCBF)$ defined on the set $\mathcal{D}$ with $\mathcal{S} \subseteq \mathcal{D} \subset{\mathbb{R}}^{n}$, if there exists an extended class $\mathcal{K}$ function ($\kappa(0) = 0$ and strictly increasing)  $\alpha$ such that
	\vspace{-0.3cm}
	\begin{equation}
	\mathop{\sup}_{u \in \mathcal{U}}[L_{f}h(x) + L_{g}h(x)u + \alpha (h(x))]\geq0, \forall x\in\mathcal{D},
	\vspace{-0.1cm}
	\end{equation}
	where $L$ represents the Lie derivatives. To be more specific:
	\vspace{-0.2cm}
	\begin{equation}
	L_{f}h(x)=\frac{\partial h(x)}{\partial x}f(x),\ L_{g}h(x)=\frac{\partial h(x)}{\partial x}g(x).
	\label{lie d}
	\vspace{-0.2cm}
	\end{equation}
	
	ZCBF is a special control barrier function that comes with asymptotic stability~\cite{Xu2015RobustnessOC}. The existence of a ZCBF implies the asymptotic stability and forward invariance of $\mathcal{S}$ as proved in~\cite{Xu2015RobustnessOC}. 
	\vspace{-0.12cm}
	\subsection{Dynamics of Quadrotor}
	\label{section:cbf-clf-qp}\vspace{-0.12cm}
	The motion of the quadrotor can be described by six degrees of freedom: The translational position $(x, y, z)$ in the inertial coordinate system $\mathbf{O}$ and attitude represented by Euler angles (roll $\phi$, pitch $\theta$, and yaw $\psi$). The vehicle attitude is defined by the rotation matrix $R\in SO(3)$~\cite{pereidaall} from the vehicle coordinate system $\mathcal{V}$ to the inertial coordinate system $\mathcal{O}$. 

	Considering the unknown wind force $d_w$ caused by wind disturbances~\cite{shi2019neural}, we adopt the following model to describe the nonlinear quadrotor dynamics,
	\vspace{-0.7cm}
	\\
	\begin{align} 
		&\dot{p} =v, \label{Za}\\ 
		& m\dot{v} = mge_3 + Re_3F+d_w, \label{uav dynamics}\\\vspace{-0.1cm}
		&\dot{R} = RS(\omega)\label{Zc}, 
	\end{align}	
		\vspace{-0.6cm}
	\\
	where $e_{3}=[0,0,1]^{T}$ is the unit vector. $m > 0$ and $g$ denote the mass and the gravity acceleration, respectively. $p=[x,y,z]^{T}$ and $v=[v_x,v_y,v_z]^{T}$ denotes the translational position and velocity in $\mathcal{O}$, respectively. $F$ denotes the total thrust generated from the four rotors in $\mathcal{V}$, and $S(.)$ is skew-symmetric mapping. The control inputs of the quadrotor are denoted by the body rotational rates $\omega = [\omega_{x}, \omega_{y}, \omega_{z}]^{T}$ and  the thrust force $F \in \mathbb{R}$ along the body axis in the vehicle coordinate system $\mathcal{V}$. $d_w = [W_{x}, W_{y}, W_{z}]^{T}$ is unknown wind force acting on the quadrotor.
	$  $
	
	In this paper, we consider the following state-space nonlinear control affine quadrotor system with partially unknown dynamics, i.e.,
	\vspace{-0.5cm}
	\\
	\begin{equation}
	\dot{q}=f(q)+G(q)u+d(q),
	\label{dynamics_with_d}\vspace{-0.2cm}
	\end{equation}
	where $q=[x,y,z,\dot{x},\dot{y},\dot{z},\phi,\theta,\psi]^T$ and $u=[F,\omega_{x},\omega_{y},\omega_{z}]^T$ denote the state and control input, respectively. $f:\mathbb{R}^{9}\rightarrow\mathbb{R}^{9}$ and $G:\mathbb{R}^{9}\rightarrow\mathbb{R}^{9 \times 4}$ compose a priori model representing our knowledge of the actual system, and $d:\mathbb{R}^{9}\rightarrow\mathbb{R}^{9}$ is the model error representing the effects of wind disturbances acting on the quadrotor.

	\subsection{Problem Statement}
	\label{subsection:problem statement}

	The safe trajectory tracking problem considered herein can be described as follows. Let the quadrotor with a limited sensing range have an initial state (consisting of initial position $x_{0} ,y_{0} ,z_{0}$, velocity $\dot{x}_{0} ,\dot{y}_{0} ,\dot{z}_{0}$, attitude $\theta_{0}$, $\phi_{0}$, $\psi_{0}$), and the reference yaw $\psi_d=0$. The quadrotor should track a given reference trajectory $p_{d} (t)=[x_d (t),y_d (t),z_d (t)]^T$ in a varying wind field while avoiding unexpected static and dynamic obstacles. If it deviates from the reference trajectory due to a sudden wind disturbance or after avoiding obstacles, it should rapidly converge to $p_{d} (t)$.

	\section{Methodology}
	\label{section:SPQC}
	
	In this section, we describe the proposed learning-based safety-preserving cascaded QP control (SPQC) for the system (\ref{dynamics_with_d}) that exploits the disturbance model to provide accurate tracking with safety and input constraints.
	
	We follow a cascaded approach described in~\cite{wu2016safety}. The cascaded controller consists of two QP controllers: a position-level QP controller and an attitude-level QP controller, as illustrated in Fig.~\ref{fig:cascaded controller}. The position-level QP controller generates the desired thrust $F$, and the attitude-level QP controller utilizes this thrust together with the high confidence uncertainty interval $D(q)$ estimated via GPs to compute the desired body rotational rates $\omega = [\omega_{x}, \omega_{y}, \omega_{z}]^{T}$. Besides, a corresponding safety constraint is designed to minimally modify each nominal QP controller to keep the quadrotor within the obstacle-free region and avoid unexpected obstacles when the quadrotor is tracking the reference trajectory.  
	\vspace{-1mm}

	\subsection{Gaussian Processes}
	
	We use the Gaussian process (GP) to learn the evolving model error $d(q)$ in (\ref{dynamics_with_d}) caused by uncertain wind disturbances $d_w$. A GP is a nonparametric regression method that can estimate complex functions and their uncertain distribution~\cite{rasmussen2006gaussian}. Given $n$ observations $\mathbf{D}_{n}:=\{q_{i},\hat{d}(q_{i})\}^{n}_{i=1}$, the mean and variance of $d(q_{*})$ at the query state $q_{*}$ can be given by
	\vspace{-1mm}
	\begin{equation}
	\mu(q_{*})=\mathbf{k}_{n}^{T}(\mathbf{K}+\sigma^{2}\mathbf{I})^{-1}\hat{\mathbf{d }}_{n},
	\label{mean}
	\vspace{-2mm}
	\end{equation}
	\begin{equation}
	\sigma^{2}(q_{*})=k(q_{*},q_{*})-\mathbf{k}_{n}^{T}(\mathbf{K}+\sigma^{2}\mathbf{I})^{-1}\mathbf{k}_{n},
	\label{var}
	\vspace{-2.0mm}
	\end{equation}
	respectively, where $\hat{\mathbf{d}}_{n}=[\hat{d}(q_{1}),\hat{d}(q_{2}),...,\hat{d}(q_{n})]$ is the observed vector subject to a zero mean Gaussian noise $\upsilon \sim\mathcal{N}(0,\sigma^{2})$. $\mathbf{K}\in\mathbb{R}^{n\times n}$ is the covariance matrix with entries, where $[\mathbf{K}]_{(i,j)}=k(q_{i},q_{j})$, $i,j \in \left\{ 1, ..., n \right\}$, and $k(q_{i},q_{j})$ is the kernel function. $\mathbf{k}_{n}=[k(q_{1},q_{*}),k(q_{2},q_{*}),...,k(q_{n},q_{*})]$, and $\mathbf{I}\in\mathbb{R}^{n\times n}$ is the identity matrix.
	
	With the system model error learned by GPs, a reliable high confidence interval  $\mathcal{D}(q)$ on the uncertain dynamics $d(q)$  can be obtained by designing the constant $c_{\delta}$~\cite{rasmussen2006gaussian}. 
	\begin{equation}
	\mathcal{D}(q)=\{d\ |\ \mu(x)-c_{\delta}\sigma(q) \leq d \leq \mu(q)+c_{\delta}\sigma(q)\}.
	\label{high confidence interval}
	\end{equation}
	
	For instance, $95.5\%$ and $99.7\%$ confidence are achieved at $c_{\delta}=2$ and $c_{\delta}=3$, respectively. 
	\subsection{Nominal Lyapunov-based Cascaded QP Controller}
	\vspace{-1mm}
	A nominal Lyapunov-based cascaded QP controller is proposed for trajectory tracking based on CLFs and GPs without the consideration of obstacle avoidance. We first construct a position-level QP controller to track the position of the quadrotor. Then, an attitude-level QP controller with a local motion planning strategy is designed to adjust the attitude of the quadrotor.
	
	\subsubsection{\textbf{Position-level QP controller}}
	Given a reference trajectory $p_{d} =[x_{d},y_{d},z_{d}]^{T}$, we can construct the following quadratic CLF:
	\vspace{-0.1cm}
	\begin{equation}
	V_{p} =\frac{1}{2}\lambda_{1}e_{p}^{T} e_{p}+\frac{1}{2}\lambda_{2}e_{\dot{p}}^{T}e_{\dot{p}},
	\label{CLF1}\vspace{-0.1cm}
	\end{equation}
	where $e_{p}=p-[x_{d},y_{d},z_{d}]^{T}$, $e_{\dot{p}}=\dot{p}-[\dot{x}_{d},\dot{y}_{d},\dot{z}_{d}]^{T}$, $p = [x,y,z]^ T$ is the current position of the quadrotor, the $\lambda_{1}>0$ and $\lambda_{2}>0$.
	
	Considering wind disturbances, GPs are employed to estimate the model uncertainties in terms of the predicted mean $\mu(q)$ and variance $\sigma(q)$ of the model error $d(q)$ through~(\ref{mean}) and~(\ref{var}). As shown by our previous work~\cite{Zheng2020LearningBasedSC}, stability constraints can be constructed based on the CLF $V_{p}$ to ensure stable high tracking performance for the uncertain dynamical system (\ref{dynamics_with_d}). Hence, based on our previous work~\cite{Zheng2020LearningBasedSC}, we can construct a position-level QP controller to compute a nominal thrust $F_n^{*}$ for tracking as follows:
	\vspace{-0.15cm}
	\begin{alignat}{2}
		\label{nominal position controller}
		F_n^{*}=&\mathop{\arg\min}_{(F,\beta) \in {\mathbb{R}^{2}}} \frac{1}{2} H_1 F + K_{\beta}\beta^{2}\\
		\mbox{s.t.} \quad
		&L_{g}V_{p}F+L_{f}V_{p}+L_{\mu}V_{p} \notag\\
		&+c_{\delta}|L_{\sigma}V_{p}|\leq - c_p V_{p}+\beta,\tag{Stability constraints} \\
		&0 \leq F \leq F_{max},\tag{Control constraints} \vspace{-0.15cm}
	\end{alignat}
	where $F_{max}\in\mathcal{U}$ is the upper bound of thrust, and $c_p$ is a positive constant. $H_1\in\mathbb{R^{+}}$, $K_{\beta}\in\mathbb{R^{+}}$, and $\beta$ is a slack variable to ensure there is no conflict among the stability constraints and control constraints in~(\ref{nominal position controller}). $L_{\mu}V_p$ and $L_{\sigma}V_p$ denote the Lie derivatives of $V_p$ with respect to $\mu$ and $\sigma$, respectively.
	
	\noindent\textbf{Remark 1}\noindent\textbf{.} Note that the optimization~(\ref{nominal position controller}) is not sensitive to the parameter $K_{\beta}$. The violation of the stability constraints can be heavily penalized as long as the $K_{\beta}$ is large enough (e.g. $10^{20}$). 
	
	\subsubsection{\textbf{Local motion planning (LMP)}}
	During trajectory tracking, a quadrotor may deviate from the reference trajectory due to a sudden wind disturbance or unexpected obstacles. In this study, we proposed an LMP algorithm to obtain a reference attitude for the quadrotor to drive it back to its reference trajectory under wind disturbances. The LMP method is described from the view of discrete control with a control period of $\Delta t$.

	Let $p(t_1)=[x(t_1),y(t_1),z(t_1)]^{T}$ and $p_{d}(t_1)=[x_{d}(t_1) ,y_{d}(t_1),z_{d}(t_1)]^{T}$ denote the position of a quadrotor and the target waypoint of reference trajectory at time $t_1$, respectively. We use GPs to estimate the mean wind disturbances $\mu_p(p(t_1))=[d_x, d_y, d_z]^{T}$ at position level, and  $\mu_{\dot{p}}(p(t_1))=[d_{\dot{x}}, d_{\dot{y}},d_{\dot{z}}]^{T}$ at velocity level based on (\ref{mean}). Hence, we can get the next nominal position $\widetilde{p}(t_2)=[\widetilde{x}(t_2),\widetilde{y}(t_2),\widetilde{z}(t_2)]^{T}$ and velocity $\dot{\widetilde{p}}(t_2)=[\dot{\widetilde{x}}(t_2),\dot{\widetilde{y}}(t_2),\dot{\widetilde{z}}(t_2)]^{T}$ of the quadrotor based on the nominal thrust $F_n^{*}$ generated by the position-level QP controller (\ref{nominal position controller}).
	
\begin{figure}[!t]
		\centering\vspace{0.1cm}
		\includegraphics[scale=0.16]{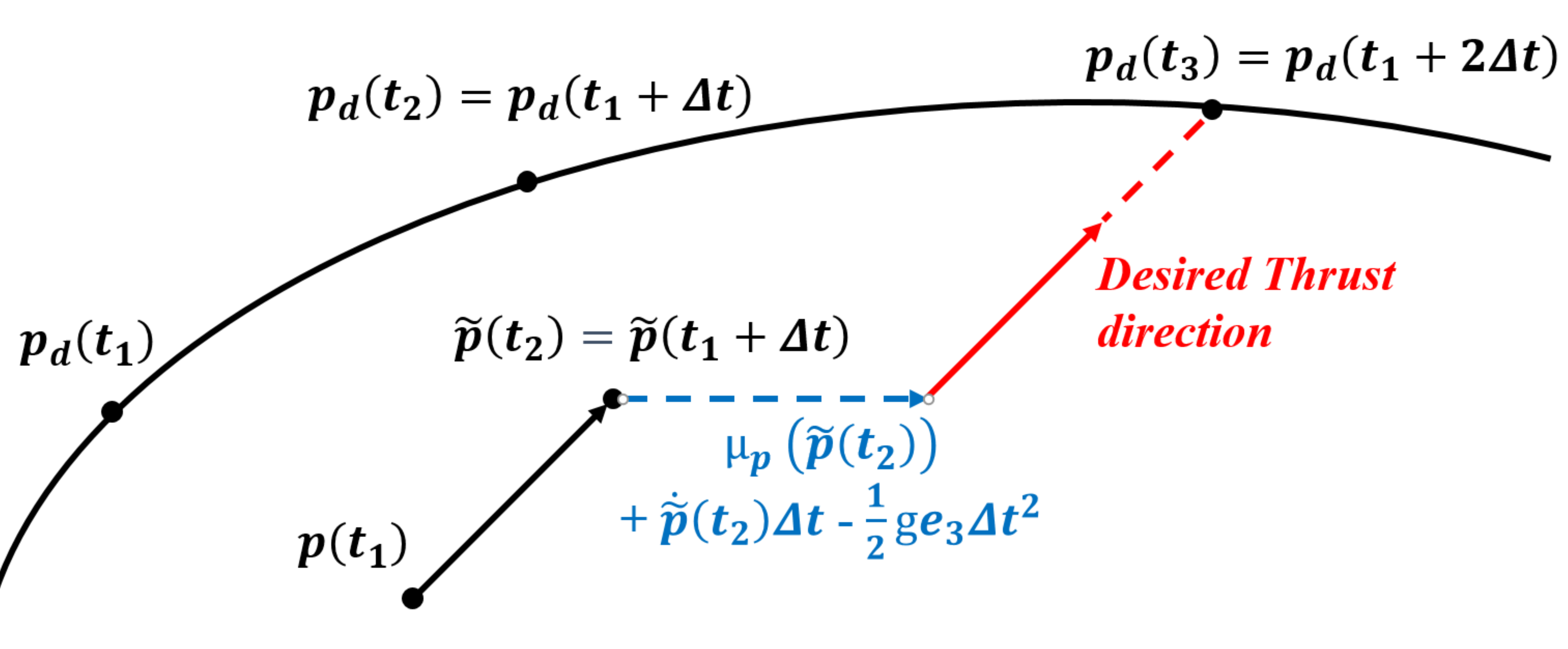}	\vspace{-0.2cm}
		\caption{{{\bf{The local motion planning algorithm.}} $p_{d}(t_1)$, $p_{d}(t_2)$ and $p_{d}(t_3)$ are three adjacent target waypoints in a reference trajectory. $\widetilde{p}(t_2)$ and $\dot{\widetilde{p}}(t_2)$ are the next nominal position and velocity, respectively. The blue line is the offset vector of the quadrotor at position level at $t_2$, and the red line is the desired thrust direction of the quadrotor at the position level.}}
		\vspace{-0.5cm}
		\label{fig:local motion planning}
	\end{figure}

	To generate the desired orientation of the quadrotor for tracking, we can adjust the orientation of thrust at time $t_1$ to obtain the desired thrust direction at position level as illustrated in Fig.~\ref{fig:local motion planning}. The desired direction of the thrust $F$ at the velocity level at $t_2$ can be obtained in a similar way to that of the desired thrust direction at the position level, and we do not show it in Fig.~\ref{fig:local motion planning} due to lack of space. 
	
	The desired direction of the thrust $F$ at position level at time $t_2$ can be denoted as follows:
	\vspace{-0.1cm}
	\begin{small} 
		\begin{equation}
		\mathbf{R_p}e_3
		=\frac{p_{d}(t_3)-\widetilde{p}(t_2)-\mu_p(\widetilde{p}(t_2))-{\dot{\widetilde{p}}(t_2)}\Delta t+\frac{ge_3}{2}(\Delta t)^2}{||p_{d}(t_3)-\widetilde{p}(t_2)-\mu_p(\widetilde{p}(t_2))-{\dot{\widetilde{p}}(t_2)}\Delta t +\frac{ge_3}{2}(\Delta t)^2||}.
		\label{position_angle}\vspace{-0.1cm}
		\end{equation}
	\end{small} 
	The desired direction of the thrust $F$ at velocity level at time $t_2$ can be denoted as follows:
	\vspace{-0.15cm}
	\begin{equation}
	\mathbf{R_v}e_3
	=\frac{\dot{p_{d}}(t_3)-{\dot{\widetilde{p}}(t_2)}-\mu_{\dot{p}}(\widetilde{p}(t_2))+ge_3\Delta t}{||\dot{p_{d}}(t_3)-{\dot{\widetilde{p}}(t_2)}-\mu_{\dot{p}}(\widetilde{p}(t_2))+ge_3\Delta t||}.
	\label{vel_angle}\vspace{-0.15cm}
	\end{equation}
	
	Then, we can obtain the desired attitude $\Omega_{pd}=[\theta_{pd},\phi_{pd},\psi_{d}]^T$ at position level and the desired attitude $\Omega_{vd}=[\theta_{vd},\phi_{vd},\psi_{d}]^ T$ at velocity level by solving~(\ref{position_angle}) and~(\ref{vel_angle}). Thus, we can obtain the desired attitude $\Omega_{d}$ as follows:
	\begin{equation}
	\Omega_{d}=\lambda_{3}\Omega_{pd} +(1-\lambda_{3})\Omega_{vd},
	\label{attitude} \vspace{-0.15cm}
	\end{equation}
	where $\lambda_{3}\in (0,1)$ is a weight coefficient.
	
	Hence, we can construct a quadratic CLF $V_{a}$ based on the desired attitude $\Omega_{d}$ as follows:
	\vspace{-0.15cm}
	\begin{equation}
	V_{a}=\frac{1}{2} \lambda_{4} e_{\Omega_{d}}^ T e_{\Omega _{d}},
	\label{CLF}\vspace{-0.15cm}
	\end{equation}
	where $e_{\Omega_d}=\Omega-\Omega_{d}$, $\Omega=[\theta,\phi,\psi]^T$, the $\lambda_{4}>0$.
	
	\subsubsection{\textbf{Orientation-level QP controller}}
	A procedure similar to design the position-level QP controller~(\ref{nominal position controller}) can be used for the formulation of the attitude-level QP controller as follows:
	\vspace{-0.45cm}
	\begin{alignat}{2}
		\label{nominal orientation controller}
		\omega_n^{*}=&\mathop{\arg\min}_{(\omega,\gamma) \in {\mathbb{R}^{4}}} \frac{1}{2}\omega^{T}H_2\omega + K_{\gamma}\gamma^{2}\\
		\mbox{s.t.} \quad
		&L_{g}V_{a}\omega+ L_{f}V_{a}+ L_{\mu}V_{a} \notag\\
		&+c_{\delta}|L_{\sigma}V_{a}| \leq - c_a V_{a}+\gamma,\tag{Stability constraints} \\
		&-\omega_{max} \leq \omega \leq \omega_{max},\tag{Control constraints}
	\end{alignat}
	\vspace{-0.5cm}
	\\
	where $\omega_{max}$ is the upper bound of body rotational rates, $H_2 \in\mathbb{R}^{3\times 3}$ is positive definite, $c_a$ is a positive constant. $\gamma$ is a slack variable for stability constraints, $K_{\gamma}\in\mathbb{R^{+}}$.
	
	\noindent\textbf{Remark 2}\noindent\textbf{.} Note that the optimization~(\ref{nominal orientation controller}) is not sensitive to the parameter $K_{\gamma}$. The violation of the stability constraints can be heavily penalized as long as the $K_{\gamma}$ is large enough (e.g. $10^{20}$).
	
	With the Lyapunov-based cascaded QP controller composed of (\ref{nominal position controller}) and (\ref{nominal orientation controller}), the quadrotor can accurately track the reference trajectory $p_d(t)$ even under wind disturbances.
	\subsection{Safety Barrier Scheme with a Limited Sensing Range}
	\vspace{-0.1cm}
	\begin{figure}[t]
		\begin{center}\vspace{0.1cm}
			\includegraphics[scale=0.19]{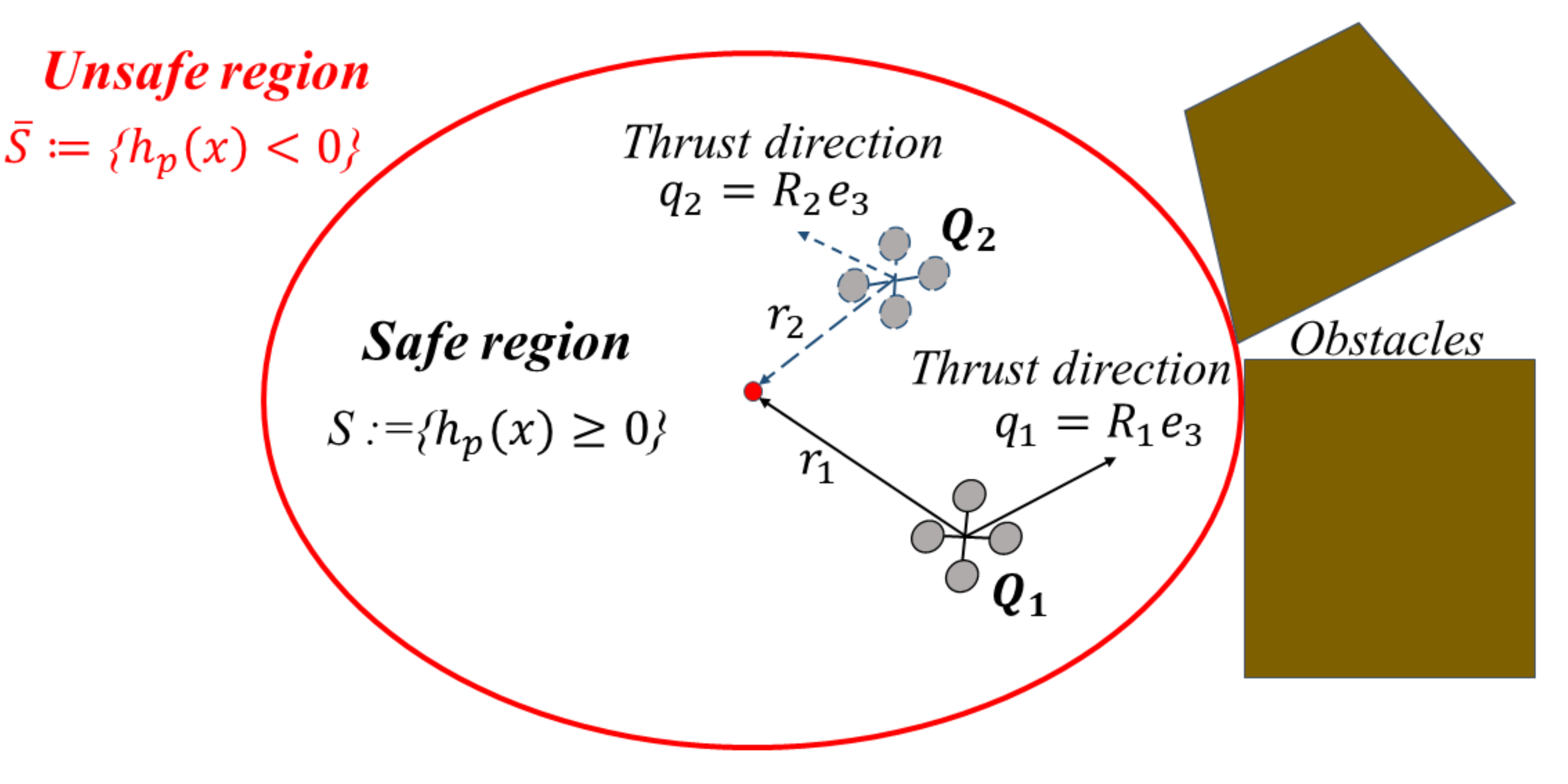}
			\vspace{-0.55cm}
		\end{center}
		\caption{{\bf{The illustration of quadrotor obstacle avoidance.}} The success of quadrotor obstacle avoidance depends both on its position and orientation. The quadrotor is in the safe state $S$ if it stays in the obstacle-free ellipsoid, and this can be captured through the condition: $h_p \geq 0$. The quadrotor Q2 is more capable of avoiding the obstacle than quadrotor Q1, and this can be captured through the condition:  $\hat{h}_R := r_{2} \cdot R_{2}e_{3} > 0 > h_{R} := r_{1} \cdot R_{1}e_{3}$.}
		\vspace{-0.5cm}
		\label{fig:safe region}
	\end{figure}

	In practice, for quadrotor tracking in cluttered environments, the reference trajectory becomes unsafe when there exist unexpected obstacles in or near the trajectory. The quadrotor should sacrifice its tracking performance and adjust its attitude to avoid unexpected obstacles as shown in Fig.~\ref{fig:safe region}. 
	
	\subsubsection{\textbf{Safety obstacle-free region}}
	An ellipsoid safe region can be obtained for the quadrotor with a limited sensing range based on the IRIS algorithm~\cite{deits2015computing}. The IRIS algorithm is efficient in quickly computing the largest ellipsoidal regions of obstacle-free space through semidefinite programming. For a quadrotor tracking task, when there exist obstacles within its sensing region, a maximum ellipsoid in mathematical form can be constructed by utilizing the distance from the quadrotor to the obstacles in each direction.
	
	\noindent\textbf{Remark 3}\noindent\textbf{.} Note that the quadrotor has a limited omnidirectional sensing range $s_r$, such that a point $\xi\in\mathbb{R}^{3\times 1}$ on an obstacle $\mathcal{B}_{s}$ is detected only when $||p-\xi|| \leq s_r$, where $p = [x,y,z]^ T$ is the current position of the quadrotor in the inertial coordinate system $\mathbf{O}$.
	
	We represent the inscribed ellipsoid as an image of the unit ball:$\varepsilon(C,\zeta) =\{Co+\zeta\ |\ \Vert o \Vert=1\}$, where $o\in\mathbb{R}^{3\times 1}$, $o^{T}o=1$ denotes a unit ball, $C\in\mathbb{R}^{3\times 3}$ denotes the mapping matrix, and $\zeta \in\mathbb{R}^{3\times 1}$ denotes the offset vector. The matrix $C$ and $\zeta$ can be obtained for specific polyhedrons (e.g. obstacles) based on the IRIS algorithm. Then, we can get the following equation via equivalent transformation:
	\vspace{-0.1cm}
	\begin{equation}
	(\varepsilon - \zeta)^{T}{C^{-1}}^{T} C^{-1}(\varepsilon - \zeta)=1.
	\label{safety set_2}\vspace{-0.1cm}
	\end{equation}
	
	Finally, we can construct a ZCBF $h_{p}$ at the position level by taking the inscribed ellipsoid as the safe region of the quadrotor:
	\vspace{-0.1cm}
	\begin{equation}
	h_{p} =1-(p- \zeta)^{T}{C^{-1}}^{T} C^{-1}(p - \zeta),
	\label{CBF1}\vspace{-0.1cm}
	\end{equation}
	where $p = [x,y,z]^ T$ is the current position of quadrotor.
	\noindent\textbf{Remark 4}\noindent\textbf{.} Note that $h_p \geq 0$ shows that the quadrotor stays within the safety ellipsoid region $S$.

	We can construct another ZCBF $h_q$ to drive the direction of the quadrotor thrust toward the center of the obstacle-free ellipsoid to avoid obstacles as illustrated in Fig.~\ref{fig:safe region}:
	\vspace{-0.15cm}
	\begin{equation}
	h_{R} =r \cdot q,	\vspace{-0.15cm}
	\end{equation}
	where $r=d -p$, $q=Re_{3}$.  
	
	\subsubsection{\textbf{Safety-preserving cascaded QP control (SPQC)}}
	A minimally modified SPQC controller can be designed to avoid obstacles based on the ZCBF $h_p$, $h_q$ and the estimated high-confidence interval $\mathcal{D}(q)$ (\ref{high confidence interval}) accounting for wind disturbances. The SPQC is designed to modify the nominal tracking controller in a way of minimally modified.
	GPs are employed to estimate the wind disturbances in terms of the predicted mean $\mu(q)$ and variance $\sigma(q)$ of the model error $d(q)$ through~(\ref{mean}) and~(\ref{var}). 
	
	As shown by our previous work~\cite{Zheng2020LearningBasedSC}, safety constraints can be enforced on the uncertain system~(\ref{dynamics_with_d}) based on the ZCBF $h_p$ to keep the quadrotor's position in the safe region $S$. Hence, the \textbf{position-level safety constraints} can be enforced as follows:
	\vspace{-0.22cm}
	\begin{algorithm}
		Position-level QP: \textit{Minimally modification}\vspace{-0.25cm}
		\begin{align}{}
		\label{Position modify}
		F^{*}=&\mathop{\arg\min}_{(F,\eta) \in {\mathbb{R}^{m+1}}} ||F-F_{n}^{*}||^2 + K_{\eta}\eta^{2}\\
		\mbox{s.t.} \quad
		&L_{f}h_p + L_{g}h_p u + L_{\mu}h_p \notag \\
		&-c_{\delta}|L_{\sigma}h_p| + \alpha_1(h_p)  \geq -\eta,\tag{Safety constraints}\\
		&F_{min} \leq F \leq F_{max},\tag{Control constraints} 
		\end{align}	\vspace{-0.5cm}
	\end{algorithm}
	\vspace{-0.32cm}
	\\
	where $F_{n}^{*}$ is the thrust generated by the nominal position-level controller~(\ref{nominal position controller}), $K_{\eta}\in\mathbb{R^{+}}$, and $\eta\in\mathbb{R}$ is a slack variable.
	
	To avoid obstacles, a safe attitude-level QP based on the ZCBF $h_R$ is designed to construct the \textbf{attitude-level safety constraints} as follows:
	\vspace{-0.22cm}
	\begin{algorithm}
		Attitude-level QP: \textit{Minimally modification}\vspace{-0.25cm}
		\begin{align}{}
		\label{Orientation modify}
		\omega^{*}=&\mathop{\arg\min}_{(\omega,\epsilon) \in {\mathbb{R}^{m+1}}} ||\omega-\omega_{n}^{*}||^2 + K_{\epsilon}\epsilon^{2}\\
		\mbox{s.t.} \quad
		&L_{f}h_R + L_{g}h_R u + L_{\mu}h_R \notag \\
		&-c_{\delta}|L_{\sigma}h_R| + \alpha_2(h_R)  \geq -\epsilon,\tag{Safety constraints}\\ 
		&-\omega_{max} \leq \omega \leq \omega_{max},\tag{Control constraints} 
		\end{align}\vspace{-0.5cm}
	\end{algorithm}
	\vspace{-0.32cm}
	\\
	where $\omega_{n}^{*}$ is the body rotational rates generated by the nominal attitude-level controller~(\ref{nominal orientation controller}), $K_{\epsilon}\in\mathbb{R^{+}}$, and $\epsilon\in\mathbb{R}$ is a slack variable.
	
	\noindent\textbf{Remark 5}\noindent\textbf{.} Note that the solution $F^{*}(x)$ and $\omega^{*}(x)$ to the QP in (\ref{Position modify}) and (\ref{Orientation modify}) are always feasible because the slack variables $\eta$ and $\epsilon$ can ensure no conflict among the safety and control input constraints.
	Furthermore, the weights $K_{\eta}$ and  $K_{\epsilon}$ are set large values (e.g. $K_{\eta}=10^{30}$, $K_{\epsilon}=10^{30}$) to penalize safety violation, and hence, the optimizations in (\ref{Position modify}) and (\ref{Orientation modify}) are not sensitive to the $K_{\eta}$ and $K_{\epsilon}$ parameters.
	
	\noindent\textbf{Remark 6}\noindent\textbf{.} Note that the safety constraints in~(\ref{Position modify}) and (\ref{Orientation modify}) are enforced to minimally modify the nominal controls generated by the Lyapunov-based cascaded QP controller composed of (\ref{nominal position controller}) and (\ref{nominal orientation controller}) via constrained QPs.

\vspace{-.04cm}
	\section{Numerical Validation}
	\vspace{-.04cm}
	\label{section:experiment}
	
	We built a simulator with Python 3.6 to numerically validate the performance of the proposed SPQC controller. The python library CVXOPT~\cite{andersen2013cvxopt} is utilized to solve the QP problem. The simulation time is set to be 20 $s$, and the control frequency is 50 $Hz$. The quadrotor model is a crazyflie 2.0 with a two-meter sensing range. The mass of the quadrotor $m = 0.027kg$ , the maximum thrust $F_{max} = 0.6N$, and the body rotational rate $|\omega| \leq 10rad\ s^{-1}$.
	
	\subsection{Experimental setup}
	
	\begin{figure}[t]		
		\centering 	\vspace{-.1cm}
		\hspace{-3mm}
		\subfigure[Tracking error]{
			\label{fig:position_error}
			\includegraphics[scale=0.06]{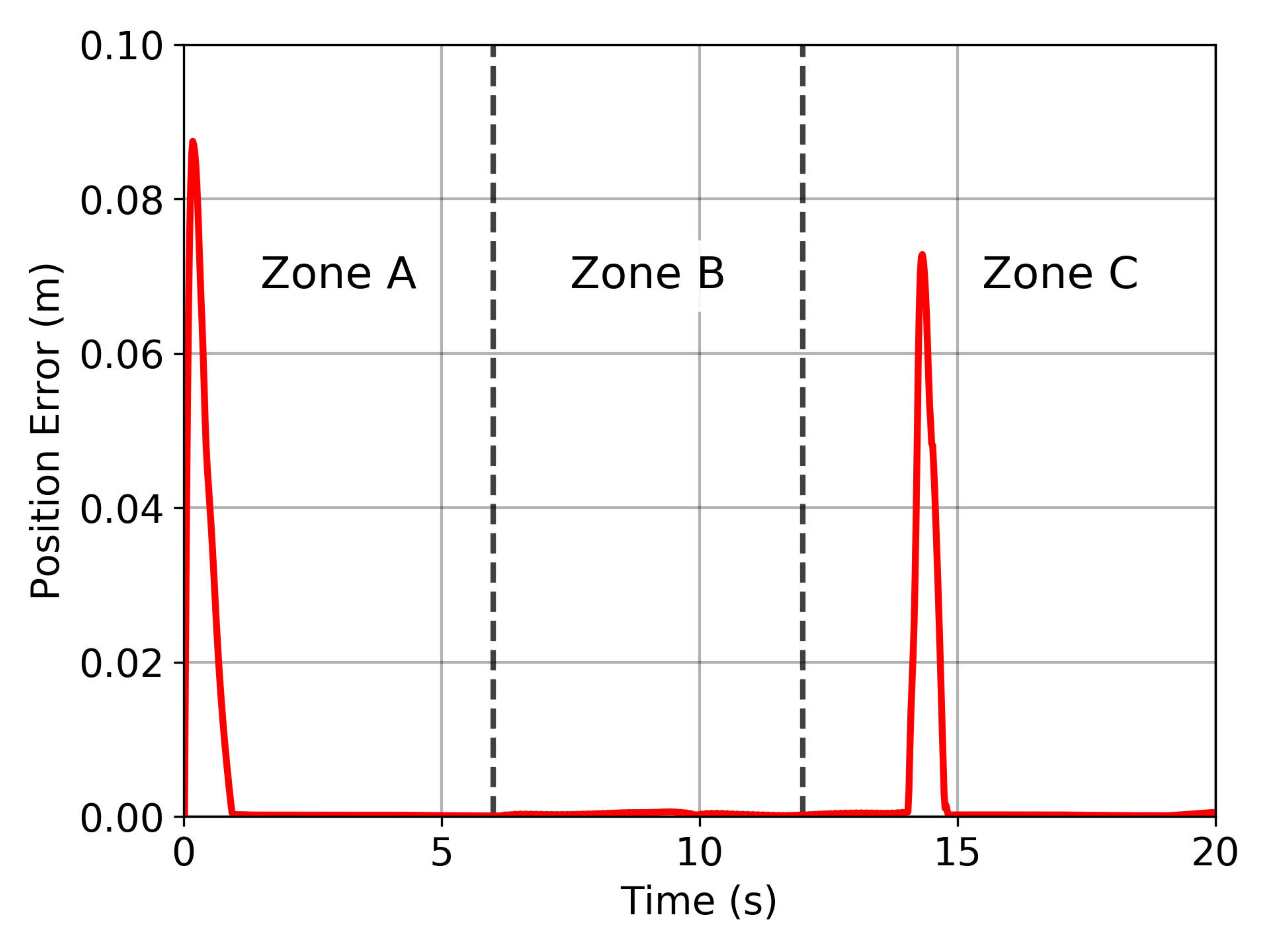}}
		\subfigure[Estimated disturbance]{
			\label{fig:estimated_disturbance}
			\includegraphics[scale=0.06]{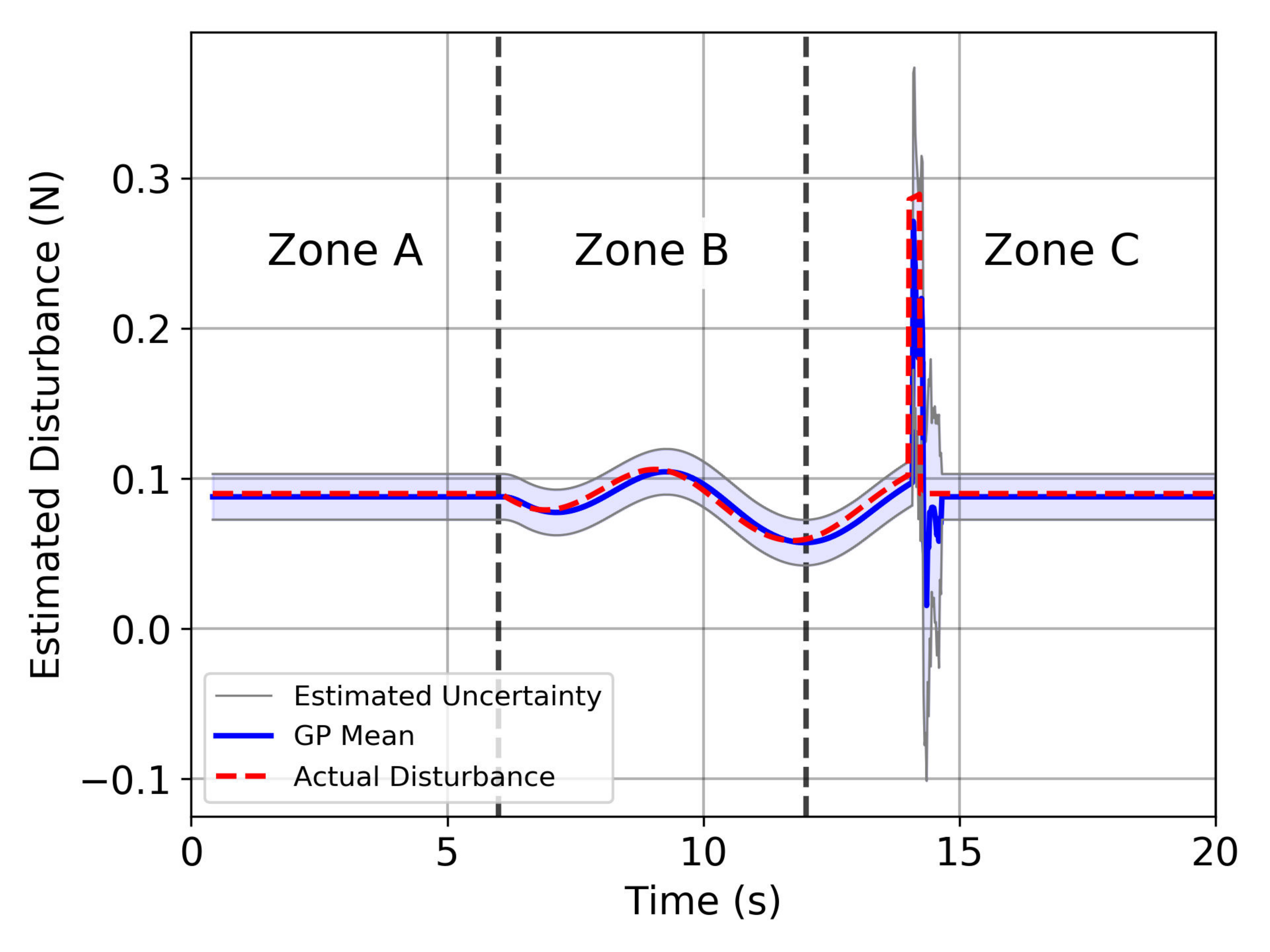}
		}  \vspace{-2mm}
		\caption{Trajectory tracking result of the SPQC controller under wind disturbances. Note that the jumps in Zone C of~\ref{fig:position_error} are caused by a sudden gust disturbance. The wind disturbance estimated via GPs is shown in~\ref{fig:estimated_disturbance}.}
		\label{fig:Trajectory tracking } \vspace{-0.4cm}
	\end{figure}
	
	\begin{table}[t]
		\renewcommand{\arraystretch}{1.1}
		\footnotesize
		\caption{Root Mean Square (RMS) Tracking Error (in meter) in Each Zone for Different Controllers on the Quadrotor}
		\label{table:rmse}
		\centering
		\vspace*{-3mm}
		\begin{tabular}{ c | c | c c c }
			\hline
			Uncertainty & Wind Disturbance &  NMPC~\cite{Owis2019QuadrotorTT} & SPQC-N &  \textbf{SPQC}\\
			\hline
			Zone A & Constant & 0.0499 & 0.0724 & 0.0198\\
			Zone B & Changing & 0.0409& 0.0883 & 0.0003\\
			Zone C & Sudden & 0.0485 & 0.0867 & 0.0137\\
			\hline
		\end{tabular}\vspace{-.6cm}
	\end{table}
	\normalsize
	
	\begin{figure*}[t]
		\centering
		\hspace{-4mm}
		\subfigure[Snapshot at t = 2s.]{
			\label{fig:A1}
			\includegraphics[scale=0.20]{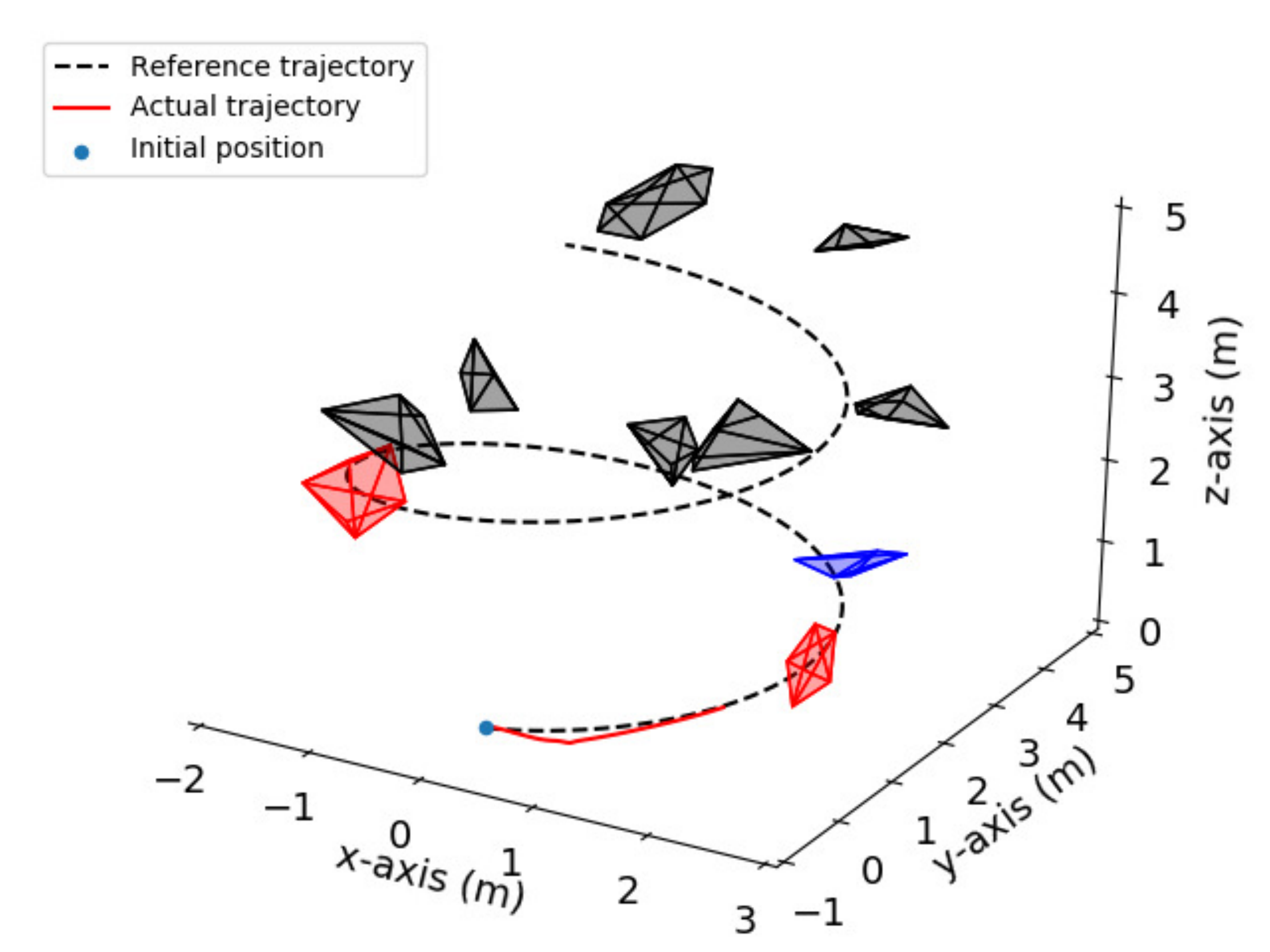}
		}\hspace{-4mm}
		\subfigure[Snapshot at t = 8s]{
			\label{fig:Vel1}
			\includegraphics[scale=0.20]{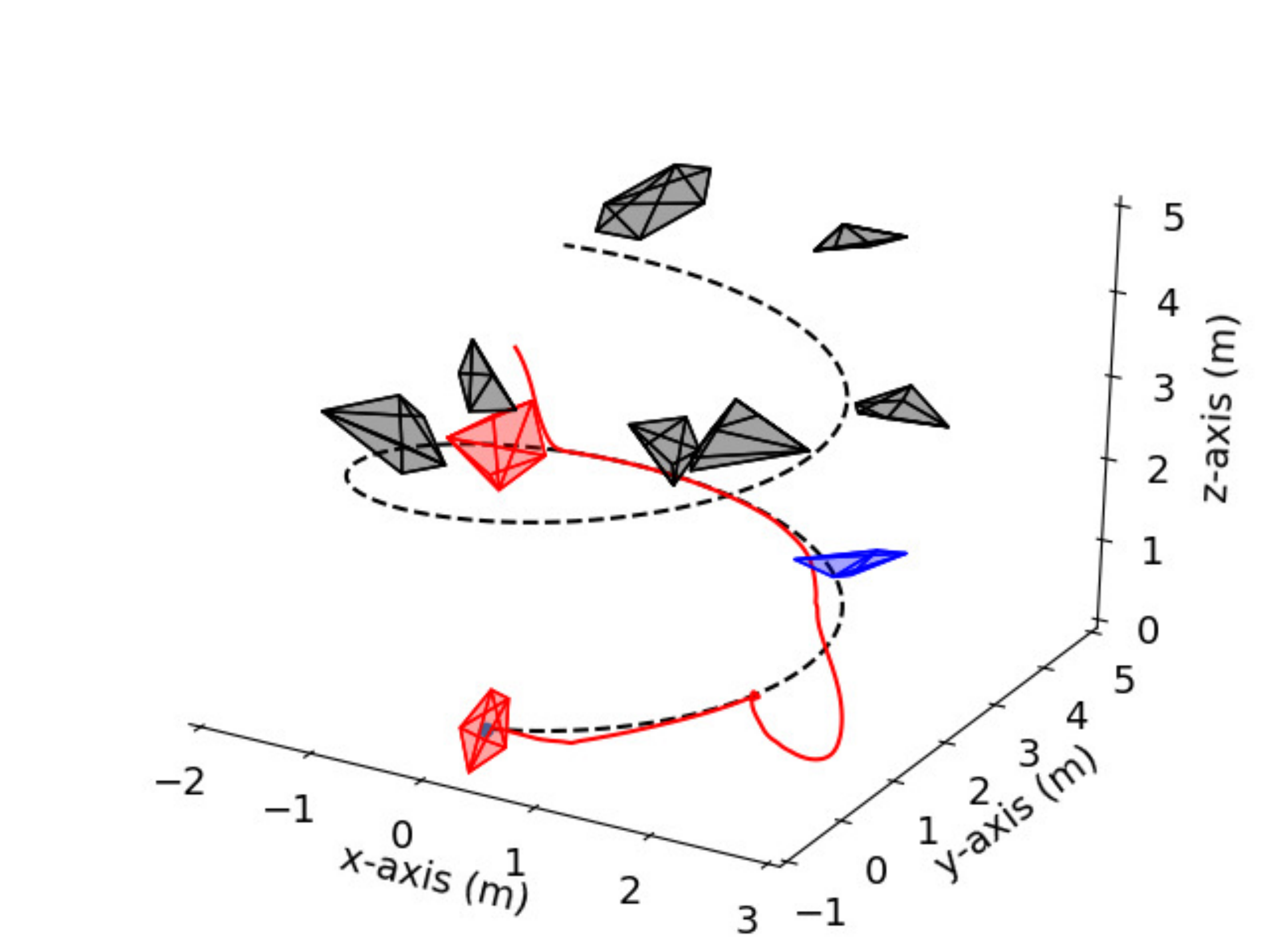}
		}\hspace{-4mm}
		\subfigure[Snapshot at t = 20s]{
			\label{fig:DB34_1}
			\includegraphics[scale=0.20]{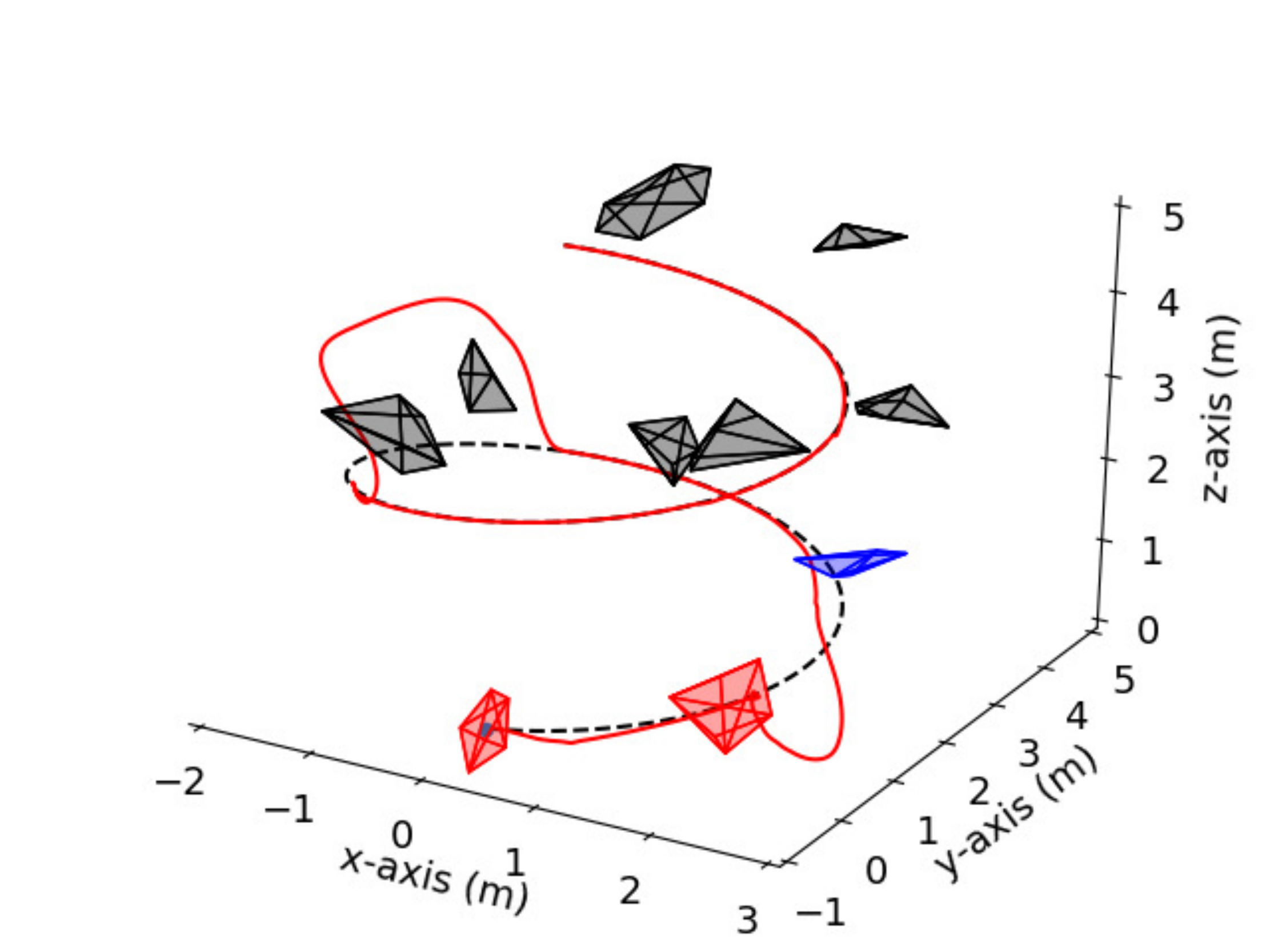}
		}\hspace{-1mm}
		\subfigure[Tracking error]{
			\label{fig:TrackingError}
			\includegraphics[scale=0.06]{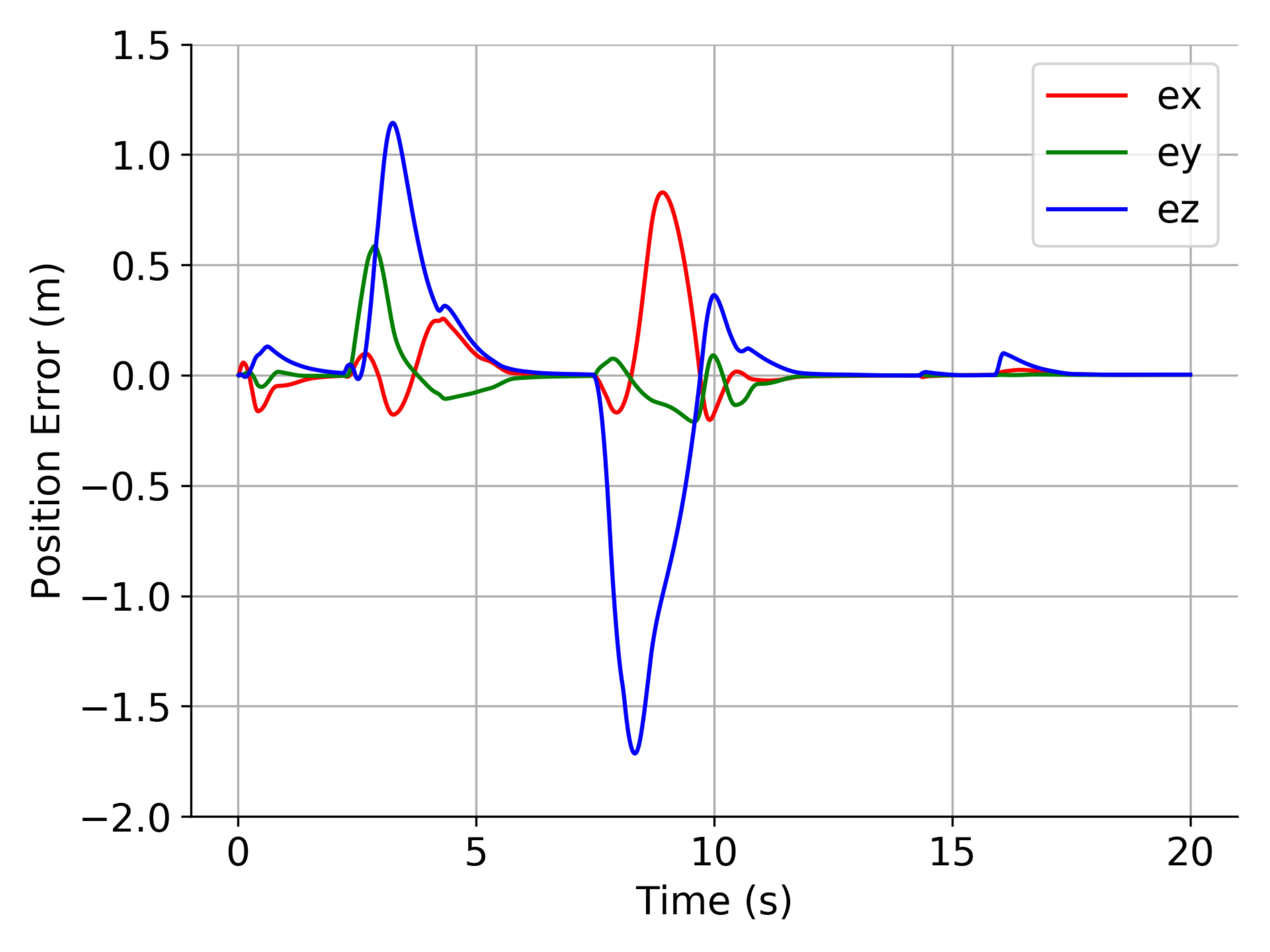}
		} \vspace{-4.5mm}	
		\caption{Numerical validation of the quadrotor flight through a densely cluttered obstacle field under varying wind disturbances. Snapshots of the simulation are shown in~\ref{fig:A1},~\ref{fig:Vel1} and~\ref{fig:DB34_1}, respectively. The red solid line and the black dashed line denote the actual and reference trajectory, respectively. The red and blue obstacles denote the dynamic and static obstacle crossing the reference trajectory, respectively. The tracking error in each axis between the actual and reference trajectory is shown in~\ref{fig:TrackingError}.When the reference trajectory violates the safety constraint (by passing through an obstacle), the controller automatically relaxes trajectory tracking to strictly enforce safety constraints. Simulation video: https://youtu.be/p-K3v9Z4OvA.}
		\label{fig:Simulation}\vspace{-6.5mm}	
		\end{figure*}	
		
	In the experiments, we assess two aspects of the proposed SPQC controller: (i) trajectory tracking performance under different wind disturbances, and (ii) trajectory tracking with obstacle avoidance under varying wind disturbances. The reference trajectory is given as a spiral curve $p_d(t)=[2sin(0.5t), 2-2cos(0.5t),0.2t]^{\top}$ and $\psi_d(t) = 0$. The initial state of the quadrotor is  $q = [\ 0\  m$, $0\ m$, $0\ m$, $0\ m/s$, $0\ m/s$, $0\ m/s$, $0\ rad$, $0\ rad$, $0\ rad$\ $]^{T}$. 
	 
	We use $3$ GPs to learn the uncertainty vector $d(q)$ (\ref{dynamics_with_d}). Each GP uses the same mixture of linear and radial basis function (RBF) kernels	$k(x,\ x^\prime)=\sigma_f^2\exp{(-\frac{1}{2}(x-x^\prime)^\mathrm{T}L^{-2}(x-x^\prime))}$  to capture the model uncertainties that result from wind disturbances, where $L=10$ and $\sigma_f=1$. Each GP uses the past $T_{s}=20$ observations collected at 50 $Hz$. To generate high confidence intervals to estimate wind disturbances, we use $[\mu(x)-3\sigma(q),\mu(x)+3\sigma(q)]$ as high confidence intervals $D(q)$ (\ref{high confidence interval}).
	
	\vspace{-.12cm}
	\subsection{Trajectory Tracking under Different Wind Disturbances\ }
	\vspace{-.12cm}
	As shown in Fig.~\ref{fig:estimated_disturbance}, we define zones A, B and C with different wind disturbances. In zone A, the wind disturbance is constant: $d_{w}=[-0.06,0.06,0.03]^{T}(N)$. In zone B, a varying wind field $\triangle d_{w} =-0.05 sin(p-p_{0})n_{w}(N)$, where $p_{0}=[0.28,4,0]^{T}(m)$ and $n_{w}=diag([0.6,-0.7,0])$ is added to $d_{w}$  to assess the adaptability to changing wind disturbances. In zone C, a large wind gust $\triangle d_{w}=[0.2, 0.18,0.1]^{T} (N)$ is added to $d_{w}$ on the system from $t=14(s)$ to $t=14.2(s)$ to push the quadrotor away from the trajectory to assess the robustness to irresistible disturbances.  
	
	It can be seen from Fig.~\ref{fig:position_error}, the position error converges to a small value in each zone. The sudden gust pushes the quadrotor away from the trajectory when the quadrotor travels in zone C, and the SPQC controller enables the quadrotor back to the trajectory after the gust. Fig.~\ref{fig:position_error} shows that the actual wind disturbance lies within the uncertainty interval estimated via GPs when the quadrotor is in zone A and B. Besides, the actual wind disturbance lies within the uncertainty interval after the sudden wind blow in zone C.
	
	We compare the SPQC controller with the NMPC~\cite{Owis2019QuadrotorTT} and an SPQC-N controller, where the SPQC-N is an ablation version of SPQC without estimated model uncertainties. The Root Mean Square Error (RMSE) of position tracking for each controller is shown in Table.~\ref{table:rmse}. In each zone, the SPQC method achieves a smaller tracking error than the NMPC and SPQC-N controller.
	\vspace{-.1cm}
	\subsection{Trajectory Tracking with Obstacles\ }
	\vspace{-.1cm}
	In the second experiment, the quadrotor is commanded to track the same dynamic time-varying trajectory $p_d(t)$ under a varying wind disturbance. The wind disturbance $d_w =[ 0.08cos(y-1), 0.08cos(x), 0.05sin(z-2)]^{T}(N)$. The reference trajectory $p_d(t)$ is surrendered by static obstacles. 
	Two dynamic obstacles fly to the quadrotor along the reference trajectory $p_d(t)$ with a speed of $0.78 m/s$ when the quadrotor is tracking the $p_d(t)$. 
	
	Figure~\ref{fig:TrackingError} shows the tracking error between the quadrotor position and the reference trajectory $p_d(t)$. When there are no obstacles along the trajectory, the tracking error converges to a small value. Furthermore, at about 2.5, 7, 14, and 16 seconds in the simulation time, the tracking errors rise to high values which indicates that the quadrotor has changed its trajectory to avoid collision with the detected obstacles. Fig.~\ref{fig:Simulation} shows that the SPQC controller stably tracks the reference trajectory when the trajectory is safe, while the controller relaxes tracking the reference when avoiding obstacles. 
	These simulation results indicate that the SPQC controller can mediate the trade-off between safety and tracking performance.
	\vspace{-.01cm}
	\section{Conclusion}
	\vspace{-.01cm}
	\label{section:conclusion}
	This paper presents a novel learning-based SPQC scheme that achieves high-accuracy tracking performance while guaranteeing safety for the quadrotor with a limited sensing range under wind disturbances. The proposed SPQC scheme consists of a nominal tracking controller and the safety constraints based on ZCBFs. In this control scheme, the GPs are exploited to estimate the uncertainties of the wind disturbances, and an LMP algorithm is designed to generate the desired tracking attitude. By minimally modifying the nominal tracking controls, the quadrotor can avoid unexpected obstacles while staying within the safety region represented by ZCBFs. The proposed SPQC algorithm was shown to improve tracking performance compared with an NMPC~\cite{Owis2019QuadrotorTT} method under different disturbances. Numerical simulation results show that the SPQC controller was capable of performing the trajectory tracking task with obstacle avoidance capacity under varying wind disturbances.
	\vspace{-2mm}
	
	
	\bibliographystyle{IEEEtran}
	\bibliography{egbib}
	\printindex
\end{document}